# Curved Trajectory Detection : A Novel Neurocognitive Perception Approach for Autonomous Smart Robots

Matin Macktoobian

*Fault Detection and Identification Lab (FDI)*
*Faculty of Electrical and Computer Engineering*
*K. N. Toosi University of Technology*
*Tehran, Iran*

matinking@hotmail.com

*Abstract* - Braitenberg vehicles could be mentioned as the seminal elements for cognitive studies in robotics fields especially neurorobotics to invent more smart robots. Motion detection of dynamic objects could be taken as one of the most inspiring abilities into account which can lead to evolve more intelligent Braitenberg vehicles. In this paper, a new neuronal circuit is established in order to detect curved movements of the objects wandering around Braitenberg vehicles. Modular structure of the novel circuit provides the opportunity to expand the model into huge sensory-biosystems. Furthermore, robust performance of the circuit against epileptic seizures is beholden to simultaneous utilization of excitatory and inhibitory stimuli in the circuit construction. Also, straight movements, as special case of curved movements could be tracked. PIONEER™, with due attention to its suitable neurosensors, is used as a Braitenberg vehicle for empirical evaluations. Simulated results and practical experiments are applied to this vehicle in order to verify new achievements of the curved trajectory detector.

## I. INTRODUCTION

One of the greatest interdisciplinary goals among scientists and researchers, which is under aegis of seminal accomplishments about nervous biosystems discovered by Cajal [1] and has syncretized both science and technology, is strengthening the intelligence abilities of the robots as much as human"s. In addition to the psychology, which could be taken as the forerunner of investigations on behavorial operation of the brain into account, recent consequences in different branches of neuroscience are led to a rather transparent viewpoint of various brain concepts. Furthermore, technological breakthroughs in bioengineering such as different types of neural networks and invention of advanced devices to track the brain conditions provide some emulated test beds in order to apply theoretical inceptions about brain operations. Based on the new achievements about role of electrical spikes in operation of the neural biosystems, it is simplicity itself that study on brain dynamics must be dedicated to dynamics of that spikes in presence of noise and the other extrinsic factors affecting the applied stimuli to the neurons, as analyezed by Hakan in [2]. Aforementioned progressions not only can improve current ideas in medicine to fix different neural seizures of the human body specially ones which could be directly referred to the construction of the brain, but will lead to effective advancements in technological stuffs like robotics to build smart robots who can think, make decisions, feel various emotions and so on like human-beings. From view point of practical implementations, Some schemes are employed to embed new neural designs into high-tech products which could be considered as efficient bed to change theoretical ideas into real sensory systems. With due attention to intrinsic switch-based nature of the neurons in human brain and their artificial fellows in artificial neural network theory, transistors are taken into account as appropriate elements to emulate neuronal systems. So, huge volume of the designed artificial neural networks which was evaluated as a serious problem for their practical realization is resolved by noticeable advances in electronics field in production of very compact integrated circuits. A successful implementation of some computational models of the neuronal circuits by VLSI technology is introduced in [3]. As previously stated, ability of building man-like machines depends on the acquired perception of the constructed robots to manipulate their environment efficiently. Valentino Braitenberg"s little creatures, as introduced in [4], could be considered as such machines which can interact with their circumferential environment by using some smart concepts. Conscious utilization of neural circuits constructed by different configurations of the neurons in order to embed various feelings and reactions stipulates that neural circuits may have the key role in development of these intelligent creatures.

Invention of Braitenberg vehicles commenced a succession of research stuffs and a great deal of strategies in various fields such as computer science, control, mechanical structures are utilized for development of these vehicles to reach more and more smart machines. Movement of Braitenberg vehicles are solely due to excitatory and inhibitory stimuli sensed by their neurosensors. Empirical attitude which usually is utilized for analyse of these creatures, could be considered as a motivation for applying fuzzy control to them, as studied in [5] and [6]. Prahitar et al tried to optimize the operation of the vehicle by genetic algorithms. [7] describes Denouements of their work against vehicle propensity for wandering beside powerful sources of stimulation. Neural networks are another smart control concept which is based on learning approaches. Combination of fuzzy and neural control, which is so-called neuro-fuzzy control, and its utilization in Braitenberg vehicles is investigated by Wang et al [8]. Furthermore, a coherent navigation strategy and its gratifying results in navigation of the robot is analysed by Yang et al in [9]. Corridor traversing might be blocked without planning new considerations which are analysed Lee et al in [10]. Neuromodulation of neural controllers which could lead to utilization of different control approaches and switching among them was another new concept described by French's researches in [11]. And as more recent achievements in territory of Braitenberg vehicles, more precise attitude toward Braitenberg vehicles and modelling of stimuli and their effects by mathematical terms is risen from Rano's works [12].

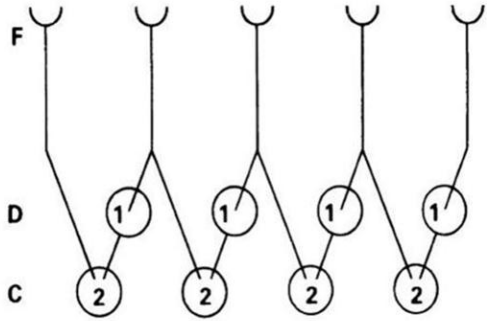

Fig 1. Braitenberg's left to right motion detector [2]

Dynamic objects wandering around Braitenberg vehicle must be tracked not only for reaching more complete and up-to-date cognition from the environment but also for making decisions about appropriate reaction regarding that motions. For example, the vehicle may establish some prudential policies when dynamic object coming close. Furthermore, motion detection neurosensors could be embedded either in visual system in cooperation with various neuron types or individually under aegis of independent proximity sensors applied to haptics system of the robot. Obviously, restricted scope of the latter approach could be considered as a blind spot for its practical realization. Straight movement detection neural circuit from left-to-right and vice versa is investigated by Braitenberg as the first prototype in this field. His proposed circuit, depicted as Fig.1, not only suffers from probable epileptic seizures due to exclusive utilization of excitatory connections but needs also the other circuit in reversed configuration corresponding to primary circuit for detection of movements in reverse direction. Hence, it will lead to increase both the number of neurons and circuit size. The other important restriction of this circuit is its poor operation in presence of curved motions. Dilatory switching of neurons in variation of frequency of input spikes prevents from in time reactions against dynamic objects.

This paper discusses a novel motion detector neural circuit with 4 main advantages in comparison with Braitenberg's prototype:

1- Collection of both left-to-right and right-to-left detectors in one circuit.
2- Modular structure of the circuit lets us use it in different sizes for detective bio-organs with various volumes.
3- Noticeable decrement in probability of epileptic seizures with combination of both excitatory and inhibitory connections in the circuit structure.
4- Dedication of 3 distinct states to the circuit corresponding to straight, near and far movements in regarding the neurosensors.

The organization of the rest of the paper is as follows, section II describes the primary straight motion detector which is introduced by Braitenberg. Presentation of curved trajectory detector in section III is classified into next 2 sections which each one is dedicated to one of the sub circuits constructing curved trajectory detector. Section VI shows the simulated and practical results. Conclusions and probable themes for future works are investigated in section VII.

## II. BRITENBERG'S MOVEMENT DETECTOR PROTOTYPE

Visual system in Braitenberg's vehicle for discerning the movements is constructed entirely of photocells. His primary 3-stage neuronal circuit for "left-to-right" detection is depicted as Fig.1.

Level (F) is responsible for receiving stimuli which are produced due to object movement. Operation of the neurosensors mounted in this level is solely relied on object existence in active scope of neurosensors. Such longer connections are in the neuronal circuit, more delay should be taken into account for that connections. With due attention to applied delay on the left branch between each adjacent pair of neurosensors in level (D), implemented neuron on right branch will be activated on time so that excitatory signals from both left and right branches will activate neuron of level (C). Left-to-right movement of the object stimulates the left-most to the right-most neurosensor one by one. Therefore, activation of neurons of level (C) regarding the direction is the same with the movement direction of the object. Obviously, neuronal circuit, which is introduced above, is not able to detect right-to-left movements. Such circuit could be acquired with displacing of delay connection and unit excitatory neuron from left branch to the right branch and vice versa, regularly. This approach not only leads to a sophisticated module but also epilepsy could be occurred with more probability. If the vehicle is surrounded by more than one dynamic object, it is possible that both sub circuits of the module become active simultaneously. In a practical point, with increasing the number of the neurons in actual systems, epileptic seizures will be evaluated as a more crucial problem.

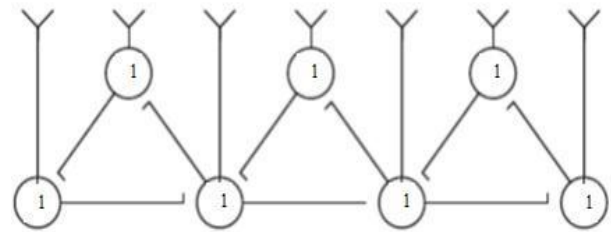

Fig 2. PDD structure

### III. CURVED TRAJECTORY DETECTOR

The novel proposed curved trajectory detector circuit is constructed from two sub circuits which are connected together serially. The first task which should be applied in operation of the curved trajectory detector, could be considered as the detection of pure direction of the movement which may be either straight or curved. For acquiring a neural circuit with high performance, Some important points from design viewpoint must be taken into account. Symmetric structure of the circuit could provide easy implementation of the circuit in various sizes and scales. One may claim that symmetric configurations could lead to superfluous utilization of seminal elements in comparison to non-symmetric ones. In one hand, scalability of the circuit could be gained by taking symmetry into account; otherwise, Variation in physical properties of the robot, its circumferential environment or dynamic objects wandering around it, will lead to noticeable sophisticated improvements in construction of the circuit. We will show that symmetric structure, which is considered for detection stuffs and is applied into this circuit, not only could be recognized as an optimized version regarding number of utilized neurons but its noticeable scalability for applying on different conditions stems from the consideration of the modularity in the circuit structure.

Furthermore, epileptic seizures always inhibits us from designing more and more fast neural circuits only based on excitatory stimuli, practically. Simultaneous activation of the neurons must be controlled in order to inhibit from crossing the secure threshold which is considered for energy consumption of the overall circuit. Epileptic seizures often happen due to crossing the local or global thresholds of the circuit. So, it is reasonable to establish efficient strategies in both sub circuits to resolve the epileptic seizures which may happen to the circuit.

### IV. PURE DIRECTION DETECTOR

With due attention to the aforementioned clarifying points for importance of application of modularity in construction of the circuit and epilepsy handling, we are going to discuss the first sub circuit of the curved trajectory detector, which is so-called „Pure Direction Detector (PDD)". The proposed structure of PDD is depicted as Fig. 2.

As shown in Fig. 2, activation threshold of the neurons in PDD is 1, which is pained on each one. As a dynamic object passes in front of the PDD, produced stimuli by mounted neurosensors are applied to the neurons. Let"s call each circular set of neurons connected with inhibitory connections as an "atomic unit". Because agglomeration of copies of this unit will construct the whole PDD. Circular inhibitory connections among every atomic unit of PDD have a key role in the prohibition of epileptic seizures; which means in every moment, just one neuron of the corresponding atomic unit could be fired and with due attention to applied relationship among that neuron and its fellows, two other neurons will be weakened within firing transition of that exaggerated neuron. Therefore, existence of each moving object in front of the vehicle will lead to exaggerate just one neuron in a PDD circuit. Furthermore, with due attention to trivial distance between each pair of neurons in each atomic unit, consideration of a bunch of stimulation sources has no bad effect on the sub circuit operation. In the case of multiple objects, there is one fired neuron in each atomic unit which is in front of the object in every moment.

Activation pattern of neurons in left-to-right (LR) and right-to-left (RL) movements are the same except in the direction of the activation. The activation pattern of both movement types for PDD sub circuit could be traced as Fig.3. Acquired sawtooth pattern stipulates that the proposed PDD is able to distinguish both LR and RL movements with reverse succession of the activation in sawtooth pattern.

### V. CURVED MOTION DETECTOR

As previously stated, PDD was utilized in order to apply a rough criterion for detection of movement direction, whereas it is unable to manifestation of neither approaching nor keeping out the robot. The second sub circuit is planned to do this task. It is reasonable to investigate on the different states of applied stimuli to the vehicle neurosensors by curved trajectory of the moving object. These different states could be shown as Fig.4.

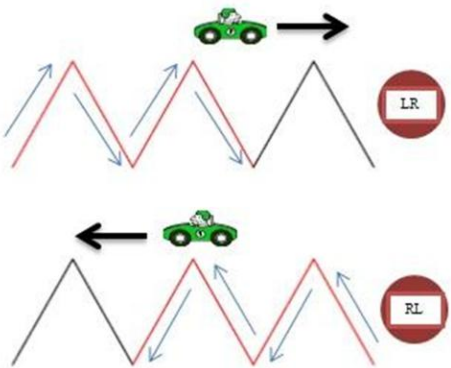

Fig 3. Traced activation pattern of PDD

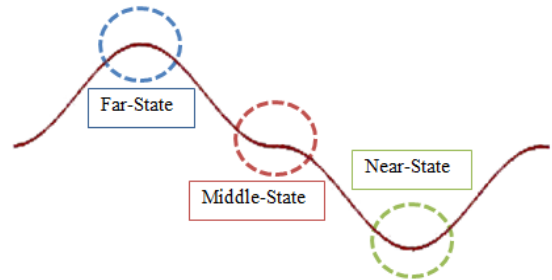

Fig 4. Traced activation pattern of PDD

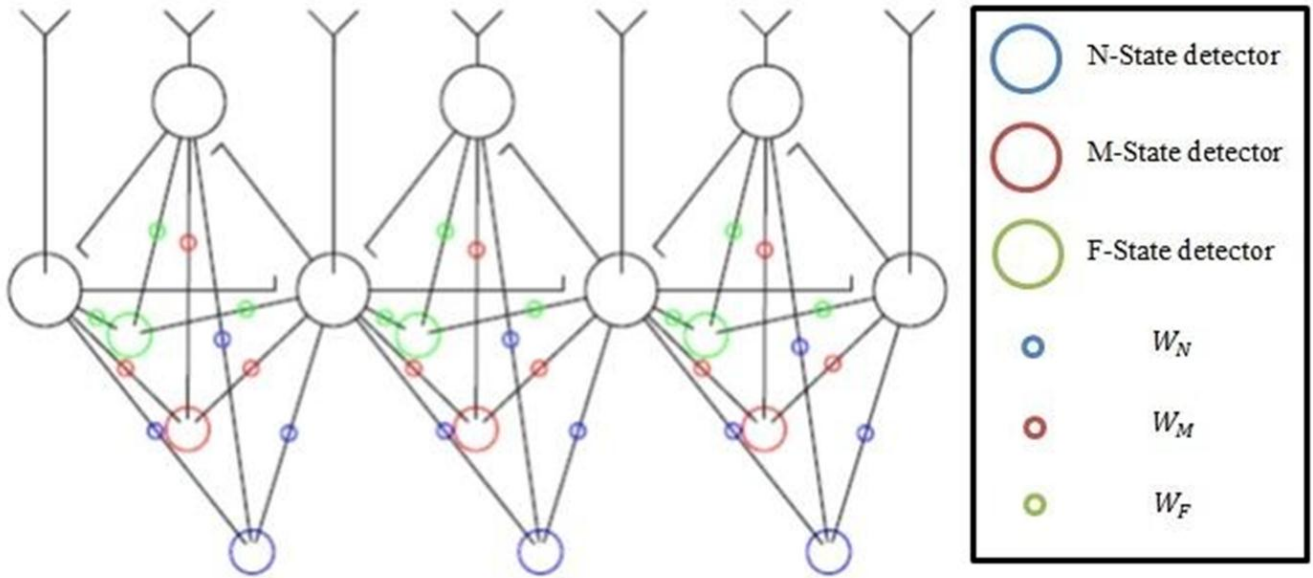

Fig 5. Curved trajectory detector

Obviously, there are 3 states which should be taken into account. Frequency of fed spikes to the neurosensor will increase as moving object is coming close. We call this state as "Near", N-state. In contrast to Near state, object may goes far and let us define "Far" state, F-state, to illustrate this case. Finally, straight trajectory of the wandering object code be coded as "Middle", M-state. Our novel sub circuit must distinguish the state of the object as one of above defined states with processing the output of PDD sub circuit as its input. Second sub circuit, which might be called „Curved motion detector (CMD)", is similar with PPD regarding utilization of an atomic unit which constructs the sub circuit. CMD atomic unit uses 3 neurons corresponding to three states, as previously stated. Each neuron will be fed by 3 excitatory connections stimulating by 3 neurons of the PDD atomic unit, whereas their thresholds should be tuned. Each output of PDD atomic unit will be manipulated by a weight unit before being applied to CMD neurons. These weights, which will be investigated carefully, let CMD fire just one of its neurons, which means the robot can detect the state of the moving object. Weights could be considered as simple gains which are applied in order to strengthen or weaken the frequency of their input spike trains. Specification of the weights must guarantee reasonable reactions of the CMD to detect the real state. If $T_i$ represents the threshold of the neuron corresponding to $i$-state, one choice might be as follows:

$$W_N < W_M < W_F$$

$$T_N < T_M < T_F$$

Which each subscript in $W_i$ terms refers to the weight specified to the $i$-state. In the sequel, with specification of appropriate values to the weights and fulfilling above inequality, when object is going far from the robot, F- state will be detected while two other neurons corresponding to N-state and F-state are off. But in the case of straight motion, not only M-state but F-state may be activated due to its planned threshold. Finally, all neurons of CMD corresponding atomic unit will be fired as robot confronts with a dynamic object approaching it. In the case of approaching to the robot this problem will be intensified. Fig. 5 shows the overall view of the curved trajectory detector circuit constructed form PDD and CMD sub circuits.

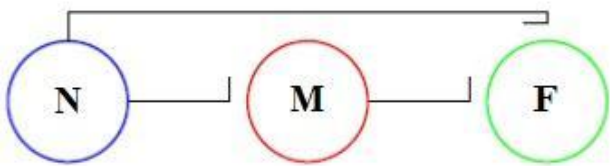

Fig 6. Inhibitory interconnection in CMD

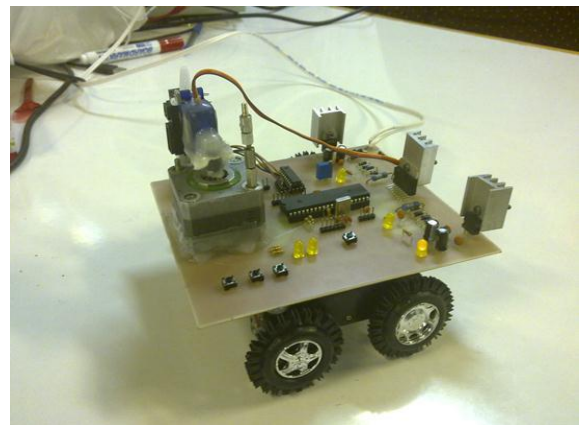

Fig 7. PIONEER™ mobile robot

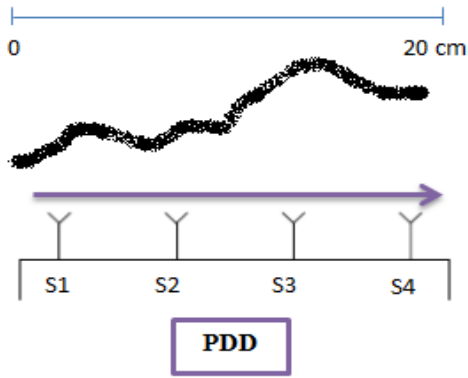

Fig 8. An empirical curved trajectory sensed by PIONEER™

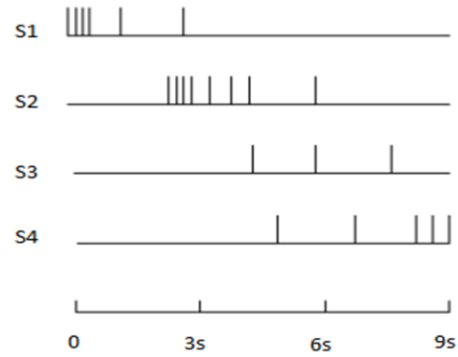

Fig 9. Monitoring of fed spikes into CTD

The detection process in two latter cases may be a simple stuff by determination of priority among fired neurons of CMD. With attention to noticeable probability of epilepsy in the case of two latter conditions specially 3$^{rd}$ one, hardware implementation of above solution could be achieved by inhibitory interconnections among neurons of CMD, which is depicted in Fig. 6.

## VI. SIMULATION DENOUEMENTS

Simulated and practical tests show noticeable consequences acquired by applying curved trajectory detector on neurorobots. PIONEER™ mobile robot, shown in Fig. 7, was utilized as a test bed for verifying theoretical ideas and the simulated results.

It is equipped with 4 IR transducers as artificial neurosensors to evaluate the performance of curved trajectory detection. Embedded IR transducers transformed it into a Braitenberg-like machine. In these experiments, the stimuli of the moving objects in front of the vehicle were modelled as spike trains which being applied in a determined range of distances from IR transducers. With due attention to desire for testing the detector under effect of different stimuli regarding intensity of the stimulation, moving objects were allowed to wander in front of the vehicle with various trajectories and produced stimulations stems from moving object are recorded and evaluates as spike trains. Such less distance till neurosensors is sensed, higher frequency of spikes will be generated. A typical curved trajectory generated by neurosensors of PIONEER™ is shown in Fig. 8. Furthermore, the scale of the distance is determined.

Stimuli are transformed into spike trains corresponding to each neurosensor which are labelled as $S_i$, $1 \leq i \leq 4$ and are shown in Fig. 9. As previously stated, PDD sub circuit is responsible for pre-processing of these spikes. Overal variation in potential level of the curved trajectory detector are shown in Fig. 10. Light variation of the potential stems from harnessing the number of fired neurons both in PDD and CMD.

## VII. CONCLUDING REMARKS AND FUTURE WORKS

Braitenberg vehicles as one of the most inspiring concepts in various investigations on neurorobotics, have a long time to be progressed in a such complete form that different neuronal and psychological operations of the brain could be applied on them. So, Robust perception of the surrounding environment could be considered as a seminal requirement for reaching to this great goal. Presented research in this paper shows that curved trajectory detection could be divided into two sub circuits. In one hand, modular architecture of sub circuits provides an appropriate scalability for using the circuit in various sizes. On the other hand, the most common crucial point in huge neuronal systems, epileptic seizure, is handled efficiently by both detector modules with simultaneous utilization of excitatory and inhibitory stimuli to ameliorate the potential level of the circuit. Tuning of weights could be more complicated with consideration of multiple moving objects and it may need to consider operational constraints and some schemes to optimize them. Also, optimized hardware implementation of neurorobotics systems like introduced curved trajectory detector on some architectures like ZISC and VLSI technology could be taken into account.

## REFERENCES

[1] Cajal, S. R. y, "Structure of the optic chiasm and general theory of the crossing visual pathways. (Structure del kiasma optico y teoria general de los entrecruzamientos de las vias nervosas.)", quoted in Cajal, 1898.

[2] H.Haken, "Brain Dynamics, An Introduction to Models and Simulations", 2$^{nd}$ Ed., Springer, Berlin 2007.

[3] H. Poor, *An Introduction to Signal Detection and Estimation*. New York: Springer-Verlag, 1985, ch. 4.

[4] V. Braitenberg, "Vehicles Experiments in synthetic psychology", The MIT Press, 1984.

[5] Beaufrere, B, "A mobile robot navigation method using a fuzzy logic approach", Robotica, No. 13, pp. 437–448, 1995.

[6] Maaref, H. and Barret, C, "Sensor-based fuzzy navigation of an autonomous mobile robot in an indoor environment", Control Eng. Pract., No. 8, pp. 747–768, 2000.

[7] Pratihar, D.K. and Dep, K., Ghosh, A, "A genetic-fuzzy approach for mobile robot navigation among moving obstacles", Int. J. Approx. Reason. 20, pp. 145–172, 1999.

[8] Wang, J.S. and Lee, C.S, "Self-adaptive recurrent neural-fuzzy control for an autonomous underwater vehicle", In Proceedings of IEEE Int. Conf. on Robotics and Automation **2**, pp. 1095–1100, 2002.

[9] X. Yang, R. V. Patel, and M. Moallem, "A Fuzzy-Braitenberg Navigation Strategy for Differential Drive Mobile Robots", Journal of Intelligent Robotic Systems, No. 47, pp. 101–124, 2006.

[10] K. Lee et al, "Topological Navigation of Mobile Robot in Corridor Environment using Sonar Sensor", Proceedings of the 2006 IEEE/RSJ International Conference on Intelligent Robots and Systems, pp. 2760-2765, 2006.

[11] R.L.B. French and L. Canamero. "Introducing neuromodulation to a Braitenberg vehicle", In Proceedings of the 2005 IEEE International Conference on Robotics and Automation, pp. 4188–4193, 2005.

[12] I. Rano, "On the Convergence of Braitenberg Vehicle 3a immersed in Parabolic Stimuli" 2011 IEEE/RSJ International Conference on Intelligent Robots and Systems, pp. 2346-2351, 2011.

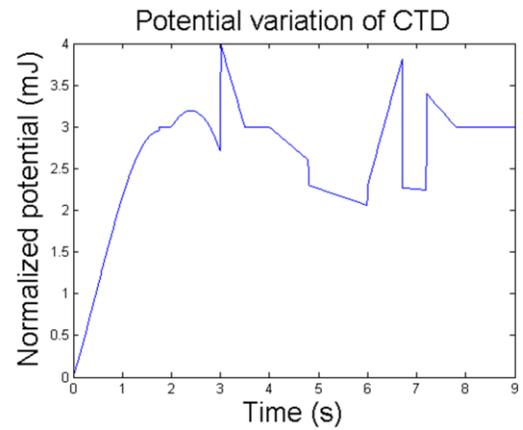

Fig 10. Potential variation of the CTD